\documentclass[letterpaper]{article} 
\usepackage{aaai19}  
\usepackage{times}  
\usepackage{helvet}  
\usepackage{courier}  
\usepackage{url}  
\usepackage{graphicx}  
\usepackage{xcolor}
\usepackage{xspace}
\usepackage{booktabs,multirow}
\usepackage{amsmath,amssymb}
\usepackage{enumitem}
\usepackage{fancyvrb}

\newcommand\singleencdec{\textsc{9enc9dec}\xspace}
\newcommand\eventPrePost{\textsc{Event2Pre/Post}\xspace}
\newcommand\eventInvol{\textsc{Event2(In)voluntary}\xspace}
\newcommand\eventXY{\textsc{Event2PersonX/Y}\xspace}
\newcommand\papername{\textsc{Atomic}\xspace}
\urldef{\atomicurl}\url{https://homes.cs.washington.edu/~msap/atomic/}

\begin{document}
%
\title{
\textsc{Atomic}: An Atlas of Machine Commonsense for \emph{If-Then} Reasoning 
}
\newcommand\aspace{\hspace{0.4em}}

\author{
Maarten Sap$^{\dagger\star}$ \aspace{} Ronan Le Bras$^\dagger$ \aspace{} Emily Allaway$^\star$ \aspace{} Chandra Bhagavatula$^\dagger$ \aspace{} Nicholas Lourie$^\dagger$ \aspace{} \vspace*{.3em} \\
{\Large\bf
Hannah Rashkin$^\star$ \aspace{} Brendan Roof$^\dagger$ \aspace{} Noah A. Smith$^{\dagger\star}$ \aspace{} Yejin Choi$^{\dagger\star}$} \vspace*{.3em}\\
$^\star$Paul G. Allen School of Computer Science \& Engineering, University of Washington, Seattle, USA\\
$^\dagger$Allen Institute for Artificial Intelligence, Seattle, USA\\
\url{msap@cs.washington.edu}
}

\newcommand{\citet}[1] {\citeauthor{#1}~\shortcite{#1}}

\maketitle
\begin{abstract}
    We present \papername, an atlas of everyday commonsense reasoning, 
    organized through 877k textual descriptions of inferential knowledge.
    Compared to existing resources that center around taxonomic
    knowledge, \papername{} focuses on inferential knowledge organized as typed \textit{if-then} relations with variables (e.g., ``\textit{if} X pays Y a compliment, \textit{then} Y will likely return the compliment''). We propose nine \textit{if-then} relation types to distinguish causes vs.~effects, agents vs. themes, voluntary vs.~involuntary events, and actions vs.~mental states.
    By generatively training on the rich inferential knowledge described in \papername{}, we show that neural models can acquire simple commonsense capabilities and reason about previously unseen events.
    Experimental results demonstrate that multitask models that incorporate the hierarchical structure of \textit{if-then} relation types lead to more accurate inference compared to models trained in isolation, as measured by both automatic and human evaluation.
\end{abstract}

\section{Introduction}
\begin{figure}[t]
    \includegraphics[width=1\columnwidth]{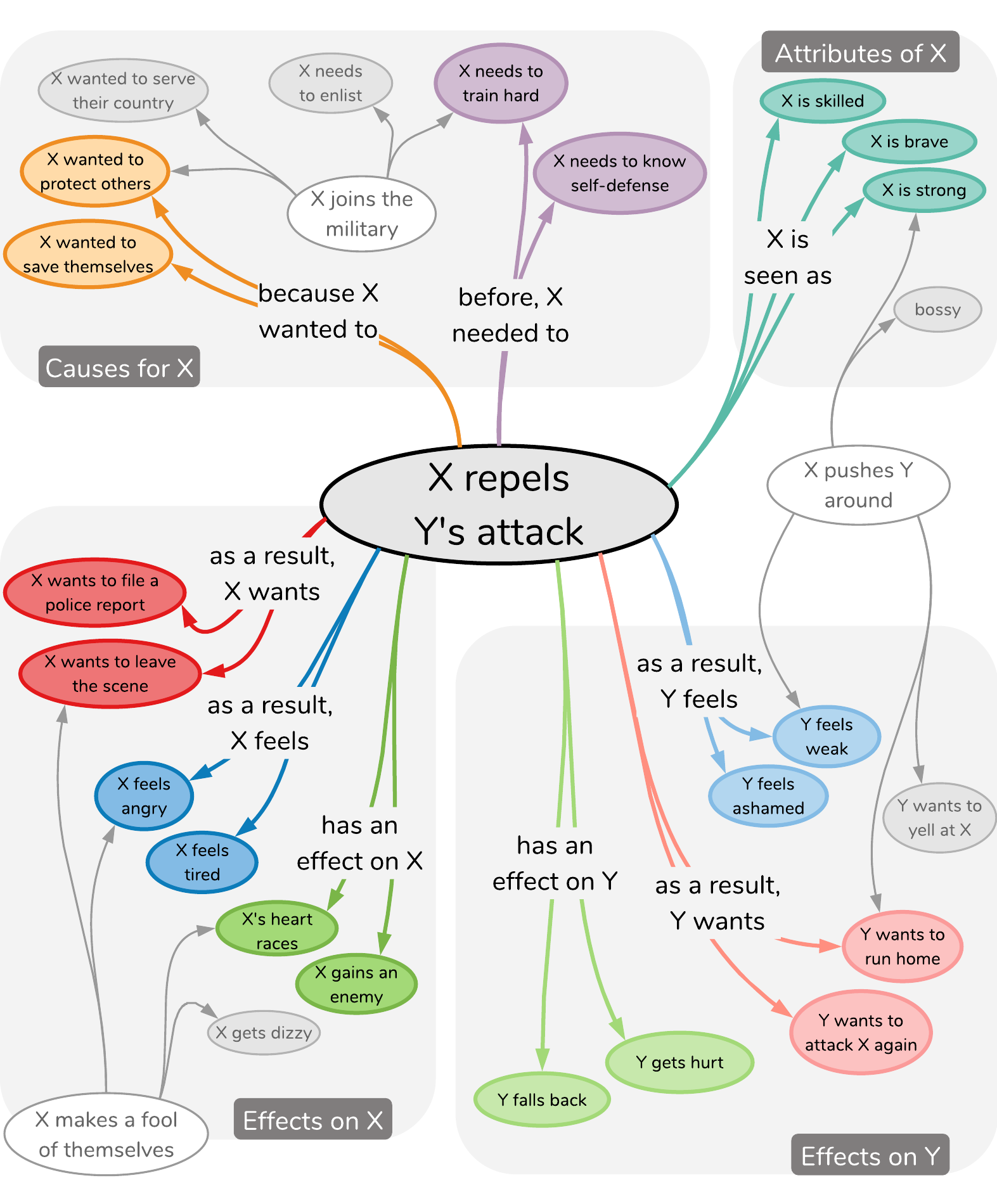}
    \caption{A tiny subset of \textsc{Atomic}, an atlas of machine commonsense for everyday events, causes, and effects.
    }
    \label{fig:intro-figure}
\end{figure}

Given a snapshot observation of an event, people can easily anticipate and reason about unobserved causes and effects in relation to the observed event: what might have happened just before, what might happen next as a result, and how different events are chained through causes and effects. For instance, if we observe an event ``X repels Y's attack'' (Figure \ref{fig:intro-figure}), we can immediately infer various plausible facts surrounding that event. In terms of the \emph{plausible motivations} behind the event, X probably wants to protect herself. As for the \emph{plausible pre-conditions} prior to the event, X may have been trained in self-defense to successfully fend off Y's attack. We can also infer the \emph{plausible characteristics} of X; she might be strong, skilled, and brave. As a \textit{result} of the event, X probably feels angry and might want to file a police report. Y, on the other hand, might feel scared of getting caught and want to run away.

\begin{figure*}[th!]
    \centering
    \includegraphics[width=0.99\textwidth]{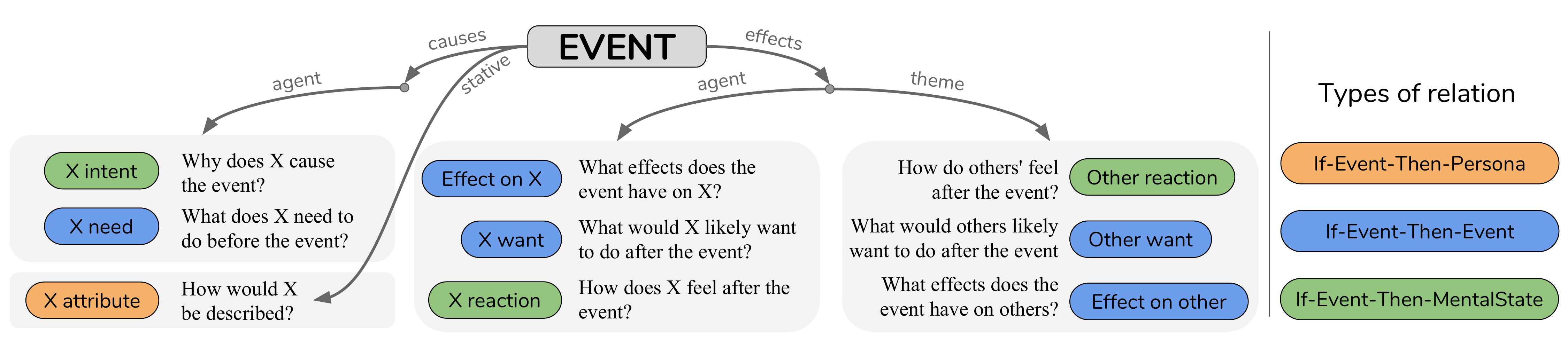}
    \caption{
    The taxonomy of \textit{if-then} reasoning types. We consider nine \emph{if-then} relations that have overlapping hierarchical structures as visualized above.
    One way to categorize the types is based on the type of content being predicted: (1) \textbf{If-Event-Then-Mental-State}, (2) 
\textbf{If-Event-Then-Event}, and (3) \textbf{If-Event-Then-Persona}.
    Another way is to categorize the types based on their causal relations: (1) \textbf{``causes''}, (2) \textbf{``effects''}, and (3) \textbf{``stative''}.
    Some of these categories can further divide depending on whether the reasoning focuses on the ``agent'' (X) or the ``theme'' (Other) of the event.
    }
    \label{fig:inference-dimensions}
\end{figure*}

The examples above illustrate how day-to-day commonsense reasoning can be operationalized through a densely connected collection of inferential knowledge. 
It is through this knowledge that we can watch a two-hour movie and understand a story that spans over several months, as we can reason about a great number of events, causes, and effects, while observing only on a small fraction of them. It also enables us to develop Theories of Mind about others \cite{moore2013development}. However, this ability, while common and trivial for humans, is    lacking in today's AI systems. This is in part because the vast majority of AI systems are trained for task-specific datasets and objectives, which lead to models that are effective at finding task-specific correlations but lack simple and explainable commonsense reasoning~\cite{Davis2015CommonsenseRA,Lake2017BuildingMT,Marcus2018DeepLA}. 

In this paper, we introduce \papername,\footnote{An \textbf{AT}las \textbf{O}f \textbf{M}ach\textbf{I}ne \textbf{C}ommonsense, available to download or browse at \scriptsize{\atomicurl}.} an atlas of machine commonsense, as a step toward addressing the rich spectrum of inferential knowledge that is crucial for automated commonsense reasoning. In contrast with previous efforts \cite{lenat1995cyc,ConceptNet} that predominantly contain taxonomic or encyclopedic knowledge \cite{Davis2015CommonsenseRA}, \papername{} focuses on inferential \emph{if-then} knowledge. The goal of our study is to create a knowledge repository that meets three requirements: scale, coverage, and quality. Therefore, we focus on crowdsourcing experiments instead of extracting commonsense from corpora, because the latter is subject to the significant reporting bias in language  that can challenge both the coverage and quality of the extracted knowledge \cite{Gordon:2013:RBK:2509558.2509563}.

We propose a new taxonomy of \emph{if-then} reasoning types as shown in Figure~\ref{fig:inference-dimensions}.
One way to categorize the types is based on the content being predicted: 
(1) \textit{If-Event-Then-Mental-State}, (2) 
\textit{If-Event-Then-Event}, and (3) \textit{If-Event-Then-Persona}.
Another way to categorize is based on their causal relations: 
(1) ``causes'', (2) ``effects'', and (3) ``stative''. 
Using this taxonomy, we gather over 877K instances of inferential knowledge. 

We then investigate neural network models that can acquire simple commonsense capabilities and reason about previously unseen events by embedding the rich inferential knowledge described in \papername.
Experimental results demonstrate that neural networks can abstract away commonsense inferential knowledge from \papername such that given a previously unseen event, they can anticipate the likely causes and effects in rich natural language descriptions.
In addition, we find that multitask models that can incorporate the hierarchical structure of if-then relation types lead to more accurate inference compared to models trained in isolation.

\section{\textit{If-Then} Relation Types}

\begin{table*}[!htb]
\footnotesize
\centering

\begin{tabular}{@{}llll@{}}
\toprule
Event                                              & Type of relations       & Inference examples                                                                                                                           & Inference dim. \\ \midrule
\multirow{7}{*}{\begin{tabular}[c]{@{}l@{}}``PersonX pays PersonY\\a compliment''\end{tabular} } & If-Event-Then-Mental-State & \begin{tabular}[c]{@{}l@{}}PersonX wanted to be nice\\ PersonX will feel good\\ PersonY will feel flattered\end{tabular}                   & \begin{tabular}[c]{@{}l@{}}xIntent\\ xReact\\ oReact\end{tabular} \\ \cmidrule(l){2-4} 
                                                   & If-Event-Then-Event       & \begin{tabular}[c]{@{}l@{}}PersonX will want to chat with PersonY\\ PersonY will smile\\ PersonY will compliment PersonX back\end{tabular} & \begin{tabular}[c]{@{}l@{}}xWant\\ oEffect\\ oWant\end{tabular} \\ \cmidrule(l){2-4} 
                                                   & If-Event-Then-Persona     & \begin{tabular}[c]{@{}l@{}}PersonX is flattering\\ PersonX is caring\end{tabular}                                                          & \begin{tabular}[c]{@{}l@{}}xAttr\\ xAttr\end{tabular}\\ \midrule
\multirow{7}{*}{\begin{tabular}[c]{@{}l@{}}``PersonX makes \\ PersonY's coffee''\end{tabular}}    & If-Event-Then-Mental-State & \begin{tabular}[c]{@{}l@{}}PersonX wanted to be helpful\\ PersonY will be appreciative\\ PersonY will be grateful\end{tabular}             & \begin{tabular}[c]{@{}l@{}}xIntent\\ oReact\\ oReact\end{tabular} \\ \cmidrule(l){2-4} 
                                                   & If-Event-Then-Event       & \begin{tabular}[c]{@{}l@{}}PersonX needs to put the coffee in the filter\\ PersonX gets thanked\\ PersonX adds cream and sugar \end{tabular}  & \begin{tabular}[c]{@{}l@{}}xNeed\\ xEffect\\ xWant\end{tabular} \\ \cmidrule(l){2-4} 
                                                   & If-Event-Then-Persona     & \begin{tabular}[c]{@{}l@{}}PersonX is helpful\\ PersonX is deferential\end{tabular}                                                        & \begin{tabular}[c]{@{}l@{}}xAttr\\ xAttr\end{tabular}\\ \midrule
\multirow{7}{*}{``PersonX calls the police''}      & If-Event-Then-Mental-State & \begin{tabular}[c]{@{}l@{}} PersonX wants to report a crime \\ Others feel worried \end{tabular}                 & \begin{tabular}[c]{@{}l@{}}xIntent\\ oReact\end{tabular} \\ \cmidrule(l){2-4} 
                                                   & If-Event-Then-Event       & \begin{tabular}[c]{@{}l@{}}PersonX needs to dial 911 \\ PersonX wants to explain everything to the police\\ PersonX starts to panic \\ Others want to dispatch some officers \end{tabular} & \begin{tabular}[c]{@{}l@{}}xNeed\\ xWant\\ xEffect \\ oWant \end{tabular} \\ \cmidrule(l){2-4} 
                                                   & If-Event-Then-Persona     & \begin{tabular}[c]{@{}l@{}}PersonX is lawful\\ PersonX is responsible\end{tabular}                                                        & \begin{tabular}[c]{@{}l@{}}xAttr\\ xAttr\end{tabular}\\ \bottomrule
\end{tabular}
\caption{\small Examples of \textbf{If-Event-Then-X} commonsense knowledge present in \papername.
For inference dimensions, ``x'' and ``o'' pertain to PersonX and others, respectively (e.g., ``xAttr'': attribute of PersonX, ``oEffect'': effect on others).}
\label{tab:annotation-examples}
\end{table*}


To enable better reasoning about events, we improve upon existing resources of commonsense knowledge by adding nine new causal and inferential dimensions. 
Shown in Figure~\ref{fig:inference-dimensions}, we define dimensions as denoting a particular type of \textit{If-Then} knowledge, answers to questions about an event, collected through crowdsourcing.
Contrary to most previous work, \papername also characterizes knowledge of events and their \textit{implied} participants (e.g., ``Alex calls for help'' implies someone will answer the call), in addition to explicitly mentioned participants (e.g., ``Alex calls Taylor for help'').

Illustrated in Table~\ref{tab:annotation-examples}, our nine dimensions span three types of \textit{If-Then} relations, outlined below.

\paragraph{If-Event-Then-Mental-State}
We define three relations relating to the mental pre- and post-conditions of an event.
Given an event (e.g., ``X compliments Y''), we reason about (i) likely \textit{intents} of the event (e.g., ``X wants to be nice''), (ii) likely \textit{(emotional) reactions} of the event's subject (``X feels good''), and (iii) likely \textit{(emotional) reactions} of others (``Y feels flattered'').

\paragraph{If-Event-Then-Event}
We also define five relations relating to events that constitute probable pre- and post-conditions of a given event.
Those relations describe events likely required to precede an event, as well as those likely to follow.
For instance, people know that ``X needs to put coffee in the filter'' before ``X makes Y's coffee''.
For post-conditions, we focus on both voluntary (``X adds cream and sugar'') and involuntary (``X gets thanked by Y'') possible next events.
We also define voluntary and involuntary possible next events for (implied) participants.

\paragraph{If-Event-Then-Persona}
In addition to pre- and post-conditions, we also define a stative relation that describes how the subject of an event is described or perceived. For instance, when ``X calls the police'', X is seen as ``lawful'' or ``responsible''.

\paragraph{An Alternative Hierarchy}
The above relation types can be categorized via a different \emph{hierarchical structure} as shown in Figure~\ref{fig:inference-dimensions}.
In particular, they can be categorized based on their causal relations: (1) ``causes'', (2) ``effects'', and (3) ``stative''. 
Each of these categories can be further divided depending on whether the reasoning focuses on the ``agent'' or the ``theme'' of the event. We omit cases where the combination is unlikely to lead to commonsense anticipation. For example, it is usually only the ``agent'' who causes the event, rather than the ``theme'', thus we do not consider that branching. We later exploit this hierarchical structure of inferential relations for designing effective neural network architectures that can learn to reason about a given event.

\section{Data}

\begin{figure}[t]
    \centering
    \includegraphics[width=\columnwidth]{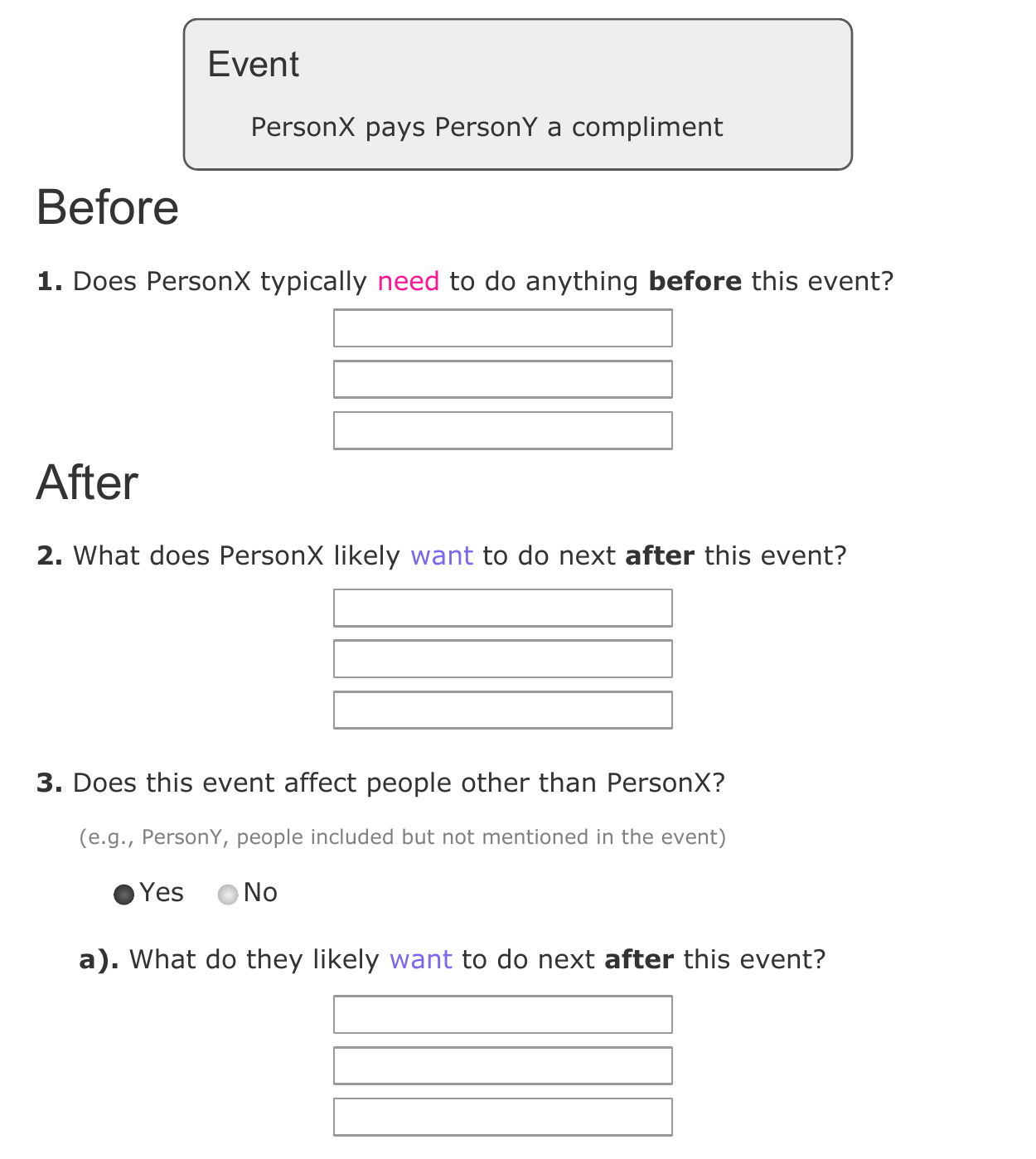}
    \caption{Template of the crowdsourcing task for gathering commonsense knowledge around events. Specific setups vary depending on the dimension annotated.}
    \label{fig:annotation-template}
\end{figure}

To build \papername, we create a crowdsourcing framework that allows for scalable, broad collection of \textit{If-Then} knowledge for given events.

\subsection{Compiling Base Events}
As \emph{base events} for our annotations, we extract 24K common event phrases from a variety of corpora. To ensure broad and diverse coverage, we compile common phrases from stories, books, Google Ngrams, and Wiktionary idioms \cite{Mostafazadeh2016corpus,Spinn3r,Ngrams}. 
Following \citet{rashkin2018event2mind}, we define events as verb phrases with a verb predicate and its arguments (``drinks dark roast in the morning'').
If a verb and its arguments do not co-occur frequently enough,\footnote{We use frequency thresholds of 5 and 100 for stories and blogs, respectively, and limit ourselves to the top 10,000 events in Google Ngrams.} we replace the arguments with a blank placeholder (``drinks \_\_\_ in the morning'').
In order to learn more general representations of events, we replace tokens referring to people with a \texttt{Person} variable (e.g. ``PersonX buys PersonY coffee''). 
In future work, other types of variables could be added for other entity references (e.g. ``PersonX moves to CityX'').

For events with multiple people explicitly involved, we run a short annotation task to help resolve coreference chains within phrases. 
Disambiguating the participants is important, since it can drastically change the meaning of the event (e.g., ``PersonX breaks PersonX's arm'' vs. ``PersonX breaks PersonY's arm'' have very different implications).
Three workers selected whether each ``Person'' mention in an event refers to PersonX, PersonY, or PersonZ, and we keep base events with combinations that at least two workers selected as valid (ppa=77\%).

\begin{table}[]
    \centering
    \begin{tabular}{l@{\hspace{15pt}}rr}
    \toprule
        & Count & \texttt{\#words}  \\ \midrule
        \# triples: If-Event-Then-* & 877,108 &  - \\
        ~~~- Mental-State & 212,598 & - \\
        ~~~- Event & 521,334 &  - \\
        ~~~- Persona & 143,176 &  - \\\midrule
        \# nodes: If-Event-Then-* & 309,515 & 2.7 \\
        ~~~- Mental-State & 51,928 & 2.1 \\
        ~~~- Event & 245,905 & 3.3 \\
        ~~~- Persona & 11,495 & 1.0  \\
        Base events & 24,313 & 4.6 \\
        \# nodes appearing $>1$ & 47,356 & -- \\
    \bottomrule
    \end{tabular}
    \caption{Statistics of \papername. Triples represent distinct \texttt{$<$event, relation, event$>$}. \texttt{\#words} represents the average number of words per node.}
    \label{tab:kg_stats}
\end{table}

\subsection{Crowdsourcing Framework}
To ensure scalability, we implement a free-form text annotation setup which asks workers to  write answers to questions about a specific event.
We chose free-text over structured or categorical annotation for two reasons.
First, categorical annotations with a large labeling space have a substantial learning curve, which limits the annotation speed and thereby the coverage of our knowledge graph.
Second, the categorical labels are likely to limit the ability to encode the vast space of commonsense knowledge and reasoning as depicted in Figure~\ref{fig:intro-figure} and Table~\ref{tab:annotation-examples}.

We create four tasks on Amazon Mechanical Turk (MTurk) (sample task in Figure~\ref{fig:annotation-template}) for gathering commonsense annotations.\footnote{The tasks were used to collect the following four sets of dimensions: (1) intent and reaction, (2) need and want, (3) effects, and (4) attributes.}\textsuperscript{, }\footnote{Our payment rate was above \$12/hour, going well beyond the federal minimum rate of \$8/hour.}
For each dimension, up to three workers are asked to provide as many as four likely annotations for an event, covering multiple possible situations (e.g., if ``PersonX drinks coffee'', then ``PersonX needed to brew coffee'' or ``PersonX needed to buy coffee''; both are distinct but likely).
Note that some events are not caused by PersonX, and some do not affect other people, making annotations for certain dimensions not necessary (specifically, for xIntent, xNeed, oReact, oEffect, and oWant) for all events.
For those dimensions, we first ask workers whether this specific inference dimension is relevant given an event.

\begin{table*}[ht!]
\centering
\begin{tabular}{@{}llc@{\hspace{12pt}}c@{\hspace{12pt}}c@{\hspace{12pt}}c@{\hspace{12pt}}c@{\hspace{12pt}}c@{\hspace{12pt}}c@{\hspace{12pt}}c@{\hspace{12pt}}c@{}}
\toprule
Dataset                         & Model                     & xIntent & xNeed & xAttr & xEffect & xReact & xWant & oEffect & oReact & oWant        \\ \midrule
\multirow{5}{*}{\textsc{Dev}}   & \singleencdec & \textbf{8.35} & 17.68 & 5.18 & \textbf{10.64} & \textbf{5.38} & \textbf{13.24} & 6.49 & 5.17 & 12.08 \\
 & NearestNeighbor & 6.14 & 11.36 & 3.57 & 5.81 & 4.37 & 7.73 & \textbf{8.02} & \textbf{6.38} & 8.94        \\ 
 \cmidrule{2-11}
 & \eventInvol  & 7.51 & \textbf{17.80} & 5.18 & 10.51 & 4.78 & 12.76 & 7.04 & 4.84 & \textbf{12.48} \\
 & \eventXY & 7.31 & 17.08 & \textbf{5.26} & 9.78 & 4.83 & 12.14 & 6.38 & 4.84 & 11.45 \\
 & \eventPrePost & 7.58 & 17.17 & -- & 10.50 & 4.73 & 11.78 & 6.71 & 4.87 & 11.52 \\ 

 \midrule
\multirow{5}{*}{\textsc{Test}}  & \singleencdec & \textbf{8.68} & 18.15 & \textbf{5.18} & \textbf{10.34} & \textbf{5.43} & \textbf{14.50} & 6.61  & 5.08 & 12.73 \\
 & NearestNeighbor & 6.64 & 11.35 & 3.37 & 5.52 & 4.59 & 8.17 & \textbf{7.58} & \textbf{5.88} & 9.18 \\
 \cmidrule{2-11}
 & \eventInvol & 7.94 & \textbf{18.22} & 5.02 & 9.78 & 4.78 & 13.67 & 7.16 & 4.71 & \textbf{13.23} \\
 & \eventXY & 7.67 & 17.33 & 5.09 & 9.45 & 4.82 & 13.19 & 6.59 & 4.68 & 11.70 \\
 & \eventPrePost & 7.96 & 17.42 & -- & 9.79 & 4.75 & 12.85 & 6.90 & 4.76 & 11.97 
         \\ \bottomrule
\end{tabular}
\caption{Average BLEU score (reported as percentages) for the top 10 generations for each inference dimension:  comparison of multitask models to single-task model. Note that BLEU scores are known to be brittle to generations worded differently from the references \cite{Liu2016howNot}. We embolden the best performing model for each dimension.} 
\label{tab:modelling-results-bleu2-devtest}
\end{table*}

\subsection{\papername Statistics}
Table~\ref{tab:kg_stats} lists descriptive statistics of our knowledge graph.
Our resulting knowledge graph contains over 300K nodes, collected using 24K base events.
Nodes in the graph are short phrases (2.7 tokens on average), ranging from 1 token for stative events (attributes) to 
3.3 and 4.6 tokens on average for more active events. 
Unlike denotational tasks where experts would only consider one label as correct, our annotations correspond to a distribution over \textit{likely} inferences~\cite{De_Marneffe2012-ia}.
To measure the degree of agreement, we run a small task asking turkers to determine whether an individual annotation provided by a different turker is valid. 
Table~\ref{tab:human-eval} shows that annotations are deemed valid on average 86.2\% of the time for a random subset of events. 
For quality control, we manually and semi-automatically detected and filtered out unreliable workers.

\section{Methods}
\newcommand{\reals}{\mathbb{R}}

Our goal is to investigate whether models can learn to perform \textit{If-Then} commonsense inference given a previously unseen event.
To this extent, we frame the problem as a conditional sequence generation problem: given an event phrase $\mathbf{e}$ and an inference dimension $c$, the model generates the target $\mathbf{t} = f_\theta(\mathbf{e},c)$.
Specifically, we explore various multitask encoder-decoder setups.

\paragraph{Encoder}
We represent the event phrase as a sequence of $n$ word vectors $\mathbf{e}=\{e_0,e_1,\ldots,e_{n-1}\} \in \reals^{n\times i_{\mathit{enc}}}$ where each word is an $i_{\mathit{enc}}$-dimensional vector.
The event sequence is compressed into a hidden representation $\mathbf{h}$ through an encoding function $f_{\mathit{enc}}: \reals^{i\times h_{\mathit{enc}}} \rightarrow \reals^h$.

In this work, we use 300-dimensional static GloVe pre-trained embeddings \cite{Pennington2014GloveGV} as our base word vectors. We augment these embeddings with 1024-dimensional ELMo pre-trained embeddings \cite{Peters2018elmo}. ELMo  provides deep contextualized representation of words using character-based representations, which allows robust representations of previously unseen events.
The encoding function is a bidirectional GRU \cite{Cho2014OnTP} of hidden size $h_{\mathit{enc}}$.

\begin{table*}[t!]
\centering
\begin{tabular}{@{}lc@{\hspace{12pt}}c@{\hspace{12pt}}c@{\hspace{12pt}}c@{\hspace{12pt}}c@{\hspace{12pt}}c@{\hspace{12pt}}c@{\hspace{12pt}}c@{\hspace{12pt}}c|c@{}}
\toprule
Model & xNeed & xIntent & xAttr & xEffect & xReact & xWant & oEffect & oReact & oWant & average \\
\midrule
\singleencdec & 48.74 & 51.70 & 52.20 & 47.52 & 63.57 & 51.56 & 22.92 & 32.92 & 35.50 & 45.32 \\
\midrule
\eventInvol & 49.82 & \textbf{61.32} & 52.58 & 46.76 & 71.22 & 52.44 & \textbf{26.46} & \textbf{36.04} & 34.70 & \textbf{47.93} \\
\eventXY & \textbf{54.04} & 53.93 & \textbf{52.98} & \textbf{48.86} & 66.42 & \textbf{54.04} & 24.72 & 33.80 & 35.08 & 46.41 \\
\eventPrePost & 47.94 & 57.77 & 52.20 & 46.78 & \textbf{72.22} & 47.94 & 26.26 & 34.48 & 35.78 & 46.76 \\
\midrule \midrule
gold \papername annotations & 81.98 & 91.37 & 78.44 & 83.92 & 95.18 & 90.90 & 84.62 & 86.13 & 83.12 & 86.18 \\
\bottomrule
\end{tabular}
\caption{Precision at 10 (\%) of generated inferences as selected by human judges for four models, averaged and broken down by dimension. We embolden the best performing model for each dimension. \eventInvol outperforms all other models significantly ($p <$ 0.05).
For comparison, we show precision of gold \papername annotations. Note that there is a varying number of gold annotations per event/dimension, while all models were constrained to make 10 predictions.}
\label{tab:human-eval}
\end{table*}

\paragraph{Decoder}
Each decoder is a unidirectional GRU of hidden size $h_{\mathit{dec}}$, with a hidden state initialized to $\mathbf{h}_{\mathit{dec}}^{(0)} = \mathbf{h}$.
The target is represented by a sequence of vectors $\mathbf{t} = \{t_0, t_1, \ldots\}$, where each $t_i \in \reals^h_{\mathit{dec}}$ is based on a learned embedding.
The decoder then maximizes $p(t_{i+1} \mid \mathbf{h}_{\mathit{dec}}^{(i)}, t_0, \ldots, t_i) =
\text{softmax}(W_o \times \text{GRU}(\mathbf{h}_{dec}^{(i)}, t_i)+b_o)$.

\paragraph{Single vs. Multitask Learning}
We experiment with various ways to combine the commonsense dimensions with multitask modeling.
We design models that exploit the hierarchical structure of the commonsense dimensions (depicted in Figure~\ref{fig:inference-dimensions}), sharing encoders for dimensions that are related.
Specifically, we explore the following models:

\begin{itemize}
    \item \eventInvol: We explore grouping dimensions together depending on whether they denote voluntary (e.g., xIntent, oWant) or involuntary (e.g., xReact, oEffect) events. This model has one encoder for four ``voluntary'' decoders, as well as another encoder for five ``involuntary'' decoders.
    \item \eventXY: We dissociate dimensions relating to the event's agent (PersonX) from those relating to the event's theme (others, or PersonY). This model has one encoder for six ``agent'' decoders as well as another encoder for three ``theme'' decoders.
    \item \eventPrePost: We split our dimensions based on whether they are related to causes (xNeed, xIntent) or effects (e.g., xWant, oEffect, xReact). In this model, there are two encoders and eight decoders.\footnote{We omit xAttr in this model, as it is trivially covered in the single task baseline.}
\end{itemize}

As a single task baseline, we train nine separate encoder-decoders, one for each dimension (\singleencdec).

\paragraph{Training Details}
To test our models, 

we split seed events into training, validation, and test sets (80\%/10\%/10\%), ensuring that events that share the same first two content words are in the same set.

As is common in generation tasks, we minimize the cross entropy of the distribution over predicted targets compared to the gold distribution in our data.\footnote{All our experiments were run using AllenNLP~\cite{allenNLP}.}
During multitask training, we average the cross entropy of each task.
Since multiple crowdworkers annotated each event, we define our training instances to be the combination of one worker's annotations.
During experiments, we use the 300-dimensional GloVe embeddings, yielding an encoder input size of $i_{\mathit{enc}}$ = 1324 once concatenated with the 1,024-dimensional ELMo embeddings.
In the encoder, ELMo's character-level modeling allows for an unlimited vocabulary.
We set the encoder and decoder hidden sizes to $h_{\mathit{enc}}=100$ and $h_{\mathit{dec}}=100$.

\section{Results}
We evaluate models on their ability to
reason about previously unseen events. Given an unseen event, models generate natural language expressions for each of the nine dimension of \textit{if-then} inferences.
We report performance using automatic scores and a human evaluation of the generated inferences.

\subsection{Automatic Scores} 
We automatically evaluate the sequence generation for each model and each inference dimension using BLEU scores.
Specifically, we compute the average BLEU score ($n = 2$, \texttt{Smoothing1}; \citeauthor{chen2014systematic}, \citeyear{chen2014systematic}) between each sequence in the top 10 predictions and the corresponding set of MTurk annotations.
As an event may not involve all nine inference dimensions (e.g., ``PersonX sees PersonX's house'' has no implications for anybody other than ``PersonX''), annotators may decide to leave an inference dimension \emph{empty}. When computing BLEU scores, we omit instances with one-third or more \emph{empty} annotations.
Table \ref{tab:modelling-results-bleu2-devtest} presents the results on both \textsc{Dev} and \textsc{Test} datasets.
The experiments show that models that exploit the hierarchical structure of the commonsense relations perform better than the model that uses separate parameters (\singleencdec).
Importantly, BLEU is a crude measure of performance as it is based on the exact match of $n$-grams and fails to capture semantically relevant generations that are worded differently~\cite{Liu2016howNot}.
As shown in Figure~\ref{fig:example-generations}, the generated samples depict varying word and phrase choices, thus we also perform human evaluation to complement automatic evaluations.

\newcommand{\robot}[0]{\includegraphics[height=10pt]{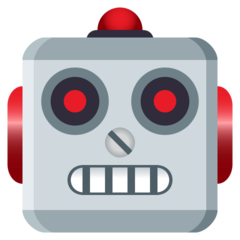}}
\newcommand{\programmer}[0]{\includegraphics[height=10pt]{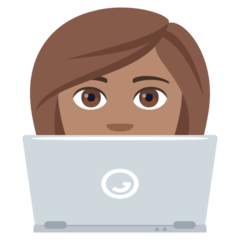}}
\begin{figure}[th!]
    \centering
    \includegraphics[width=\columnwidth]{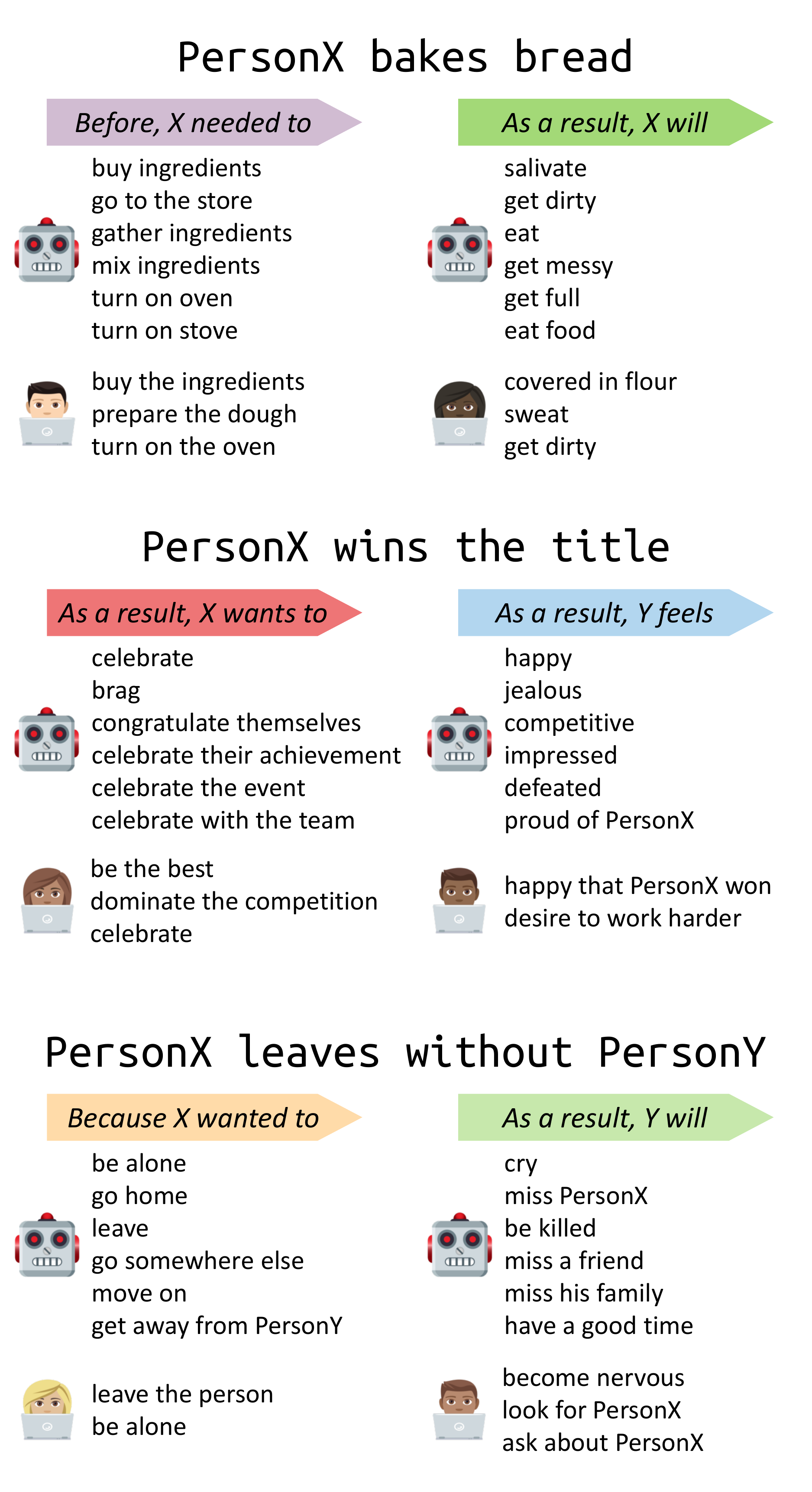}
    \caption{Examples of machine (\robot) generated inferences for three events from the development set, ordered from most likely (top) to least likely (bottom) according to the \eventInvol model. Human (\programmer) generated inferences are also shown for comparison.}
    \label{fig:example-generations}
\end{figure}

\subsection{Human Evaluation}

Since automatic evaluation of generated language is an open research question~\cite{Liu2016howNot}, we also assess our models' performance through human evaluation.
We randomly select 100 events from the test set and use beam search to generate the 10 most likely inferences per dimension.
We present five crowdworkers with the 10 generated inferences, and ask them to select all inferences they think are valid.
Table~\ref{tab:human-eval} shows each model's precision at 10, computed as the average number of correct generations per dimension.
Following the same crowdsourcing setup, we also assess the quality of the gold \papername annotations for the same set of test events. Human evaluation (last line of Table~\ref{tab:human-eval}) indicates that 86.2\% of the descriptions are valid, 
showcasing the quality of commonsense knowledge contained in \papername.

Human evaluation supports our conclusion from automatic evaluation -- that models that leverage the \textit{if-then} hierarchy perform better than models that don't.
Specifically, explicitly modeling whether inference dimensions describe voluntary actions (e.g., what X wants to do next) or involuntary effects (e.g., X or Y's reactions) yields more sensible generations, as evidenced by the performance of \eventInvol.

\subsection{Qualitative Results}
We present sample commonsense predictions in 
Figure~\ref{fig:example-generations}. Given an event ``PersonX bakes bread'', our model can correctly infer that X probably \emph{needs} to ``go to the store'' or ``mix ingredients'' or ``turn on the oven''. Our model also correctly predicts that the likely effect of this event would be that X will ``get dirty'' or ``eat food''.

\subsection{Comparison with ConceptNet}

ConceptNet \cite{speer2017conceptnet} represents commonsense knowledge as a graph of \textit{concepts} connected by \textit{relations}. Concepts consist of words or phrases, while relations come from a fixed set of edge types.

While ConceptNet captures general commonsense knowledge---much of which is taxonomic in nature\footnote{While ConceptNet includes various inferential  relations  (e.g.,  “entails”, “causes”, “motivated by”), their instances amount to only about 1\% of ConceptNet.}---\papername{} focuses on sequences of events and the social commonsense relating to them. This focus means that while events and dimensions in \papername{} loosely correspond to concepts and relations from ConceptNet, individual dimensions, such as \textit{intents}, can't be mapped cleanly onto any combination of ConceptNet's relations. The correspondence is neither one-to-one nor one-to-many. Still, in order to empirically investigate the differences between ConceptNet and \papername{}, we used the following best-effort mappings between the dimensions and relations:

\begin{itemize}[itemsep=0em]
    \item \textbf{Wants}: \textsc{MotivatedByGoal}, \textsc{HasSubevent}, \textsc{HasFirstSubevent}, \textsc{CausesDesire}
    \item \textbf{Effects}: \textsc{Causes}, \textsc{HasSubevent}, \textsc{HasFirstSubevent}, \textsc{HasLastSubevent}
    \item \textbf{Needs}: \textsc{MotivatedByGoal}, \textsc{Entails}, \textsc{HasPrerequisite}
    \item \textbf{Intents}: \textsc{MotivatedByGoal}, \textsc{CausesDesire}, \textsc{HasSubevent}, \textsc{HasFirstSubevent}
    \item \textbf{Reactions}: \textsc{Causes}, \textsc{HasLastSubevent}, \textsc{HasSubevent}
    \item \textbf{Attributes}: \textsc{HasProperty}
\end{itemize}

\noindent
We then computed the overlap of \texttt{<event1, dimension, event2>} triples in \papername{} with the \texttt{<concept1, relation, concept2>} triples in ConceptNet. 
We found the overlap to only be as high as 7\% for \textit{wants}, 6\% for \textit{effects}, 6\% for \textit{needs}, 5\% for \textit{intents}, 2\% for reactions, and 0\% for attributes. Moreover, only 25\% of the events in \papername{} are found in ConceptNet. 
Thus, \papername{} 
offers a substantial amount of new inferential knowledge that has not been captured by existing resources.

\section{Related Work}

\paragraph{Descriptive Knowledge from Crowdsourcing}
Knowledge acquisition and representation have been extensively studied in prior research~\cite{OpenMindCommonsense,ConceptNet,lenat1995cyc}.
However, most prior efforts focused on taxonomic or encyclopedic knowledge \cite{Davis2015CommonsenseRA}, which, in terms of epistemology, corresponds to \emph{knowledge of ``what''}.
Relatively less progress has been made on \emph{knowledge of ``how''} and \emph{``why''}.
For example, OpenCyc 4.0 is a large commonsense knowledge base consisting of 239,000 concepts and 2,039,000 facts in LISP-style logic~\cite{lenat1995cyc},
known to be mostly taxonomic~\cite{Davis2015CommonsenseRA}.
In fact, only 0.42\% of \papername events appear in OpenCyc, which we found contains 99.8\% relations that are either taxonomic (\verb|isA|), string formatting relations, or various definitional relations. A typical example is shown below:
\begin{footnotesize}
\begin{verbatim}
(genls (LeftObjectOfPairFn
   SuperiorLobeOfLung) LeftObject)
(isa (WordNetSynsetReifiedFn
   460174) WordNetSynset)
(genls (AssociatesDegreeInFn
   EngineeringField) AssociatesDegree)
\end{verbatim}
\end{footnotesize}

\noindent
Importantly, these LISP-based representations of OpenCyc are non-trivial to integrate into modern neural network based models, as it is not straightforward to compute their embedding representations. In contrast, the natural language representations in \papername can be readily used to obtain their neural embeddings, which can also be mixed with pretrained embeddings of words or language models.

Similarly,  ConceptNet \cite{speer2017conceptnet} represents commonsense knowledge as a graph that connects words and phrases (\textit{concepts}) with labeled edges (\textit{relations}).
While ConceptNet provides relatively more inferential relations (e.g., ``entails'', ``causes'', ``motivated by''), they still amount to only about 1\% of all triples in the graph. 
In contrast, \papername{} is centered around events represented with natural language descriptions.
While events and dimensions in \papername{} loosely correspond to concepts and relations in ConceptNet, the two represent very different information and ultimately have relatively small overlap as discussed in the Results section.

Recent work by \citet{gordon2017formal} compiles a list of nearly 1,400 commonsense axioms in formal logic, which connect abstract concepts to each other. For example, they define an \verb|event| as being made up of \verb|subevents|, expressed by:
\begin{small}
\begin{verbatim}
(forall (e)
 (iff (event e)
  (or (exists (e1 e2) 
   (and (nequal e1 e2)(change' e e1 e2)))
    (exists (e1)
     (subevent e1 e)))))
\end{verbatim}
\end{small}
These axioms are abstract in that they are not grounded with respect to specific objects, events, or actions.  In contrast, our work presents 880K triples of commonsense knowledge expressed in natural language and fully grounded with concrete events, actions, mental states.

The recent work of \citet{rashkin2018event2mind} introduced a commonsense inference task about events and mental states: given an event described in natural language, the task is to generate the reaction and intent of actors involved in the event. \papername{} is inspired by this work, but substantially scales up (i) the crowdsourcing procedure to nine dimensions per event, and (ii) the size of the knowledge graph---from 77K events in Event2Mind to 300K events in \papername{}. Moreover, while the primary focus of \cite{rashkin2018event2mind} was inferential knowledge, its scope was limited to mental states.

\paragraph{Acquired Knowledge from Extraction and Induction}
More generally, the goal of moving beyond static commonsense knowledge to enable automated commonsense reasoning has inspired much research. Several projects have sought to extract commonsense inferential rules from naturally occurring resources such as large corpora \cite{Schubert:2002:WDG:1289189.1289263}, movie scripts \cite{Tandon2017WebChild2}, and web how-tos \cite{Chu2017DistillingTK}. Such systems must inevitably deal with reporting bias \cite{Gordon:2013:RBK:2509558.2509563}, or the fact that the frequency and selection of phenomena represented in natural language systematically differ from what occurs in the real world. Other approaches have sought to induce commonsense rules from large knowledge bases \cite{Galrraga2013AMIEAR,Yang2015Embedding}. While these approaches have also had success, the choice of schema and information represented in current knowledge bases limits the scope of propositions such systems can learn.

\paragraph{Scripts and Narrative Reasoning}
Other work has focused more specifically on representing and reasoning about sequences of events, similarly to \papername.
Early work on event sequences studied \textit{scripts}, a kind of structured representation for prototypical sequences of events \cite{schank1977scripts}.
More recently, \textit{narrative event chains} have been proposed as a similar formalism for prototypical sequences of events that may be learned from raw text \cite{Chambers2008UnsupervisedLO}. 
This work additionally proposed the \textit{Narrative Cloze Test} as a benchmark for story understanding. In contrast to narrative event chains, the \textit{ROC Stories Corpus} crowdsources event sequences represented as natural language stories rather than using a specific formalism \cite{Mostafazadeh2016corpus}.
Additionally, the \textit{Story Cloze Test} adapts these stories into a new benchmark by requiring systems to choose between the true and a false ending to the story.
Our work interpolates between these two approaches by representing events in natural language while structuring the relationships between events into the edges of a graph.
The \textit{Choice of Plausible Alternatives} (COPA) task offers a similar benchmark for commonsense understanding of events and their relationships \cite{Roemmele2011ChoiceOP}.
In COPA, a system is presented a premise and two alternatives that might have a causal relationship with the premise. While COPA, like \papername{}, represents events as free-form text with structured relationships, it covers only a limited number of relations (cause and effect) and is smaller in scale (contains only 1,000 instances).

\section{Conclusion}
We present \papername{}, an atlas of everyday commonsense inferential knowledge about events described in natural language and associated with typed \textit{if-then} relations. \papername{} consists of over 300k events associated with 877k inferential relations,
making it the largest knowledge graph of its kind. Our crowdsourcing framework gathers annotations in the form of free-form textual responses to simple questions which enables large-scale high quality collection of commonsense about events. We also present neural network models that can learn to reason about previously unseen events to generate their likely causes and effects in natural language. 

\section*{Acknowledgments}
We thank the anonymous reviewers for their many insightful comments. We also thank Peter Clark, Dan Weld, Keisuke Sakaguchi, Vidur Joshi, Mark Neumann, xlab, Mosaic and AllenNLP team members, for their helpful comments and suggestions. 
Experiments were conducted on the AllenAI Beaker platform.
This work was supported in part by
NSF GRFP DGE-1256082, NSF IIS-1714566, IIS-1524371, IIS-1703166, Samsung AI Grant, DARPA CwC program through ARO (W911NF-15-1-0543), and the IARPA DIVA program through D17PC00343.

\bibliographystyle{aaai}
\bibliography{atomic}

\end{document}